\documentclass[10pt,twocolumn,letterpaper]{article}

\usepackage[pagenumbers]{cvpr} % To force page numbers, e.g. for an arXiv version

\definecolor{cvprblue}{rgb}{0.21,0.49,0.74}
\usepackage[pagebackref,breaklinks,colorlinks,allcolors=cvprblue]{hyperref}
\usepackage{bm}
\usepackage{multirow}
\usepackage{bbding}

\def\paperID{****} % *** Enter the Paper ID here
\def\confName{****}
\def\confYear{****}

\title{Proc-GS: Procedural Building Generation for City Assembly with 3D Gaussians}

\author{Yixuan Li$^1$, Xingjian Ran$^2$, Linning Xu$^1$, Tao Lu$^3$, Mulin Yu$^2$, Zhenzhi Wang$^1$ \\ Yuanbo Xiangli$^4$, Dahua Lin$^{1,2}$, Bo Dai$^{5,2}$\Envelope\\
	$^1$ The Chinese University of Hong Kong \quad
	$^2$ Shanghai Artificial Intelligence Laboratory \\
        $^3$ Brown University \quad
        $^4$ Cornell University \quad
        $^5$ The University of Hong Kong \\
}

\begin{document}
% \maketitle

\twocolumn[{
    \renewcommand\twocolumn[1][]{#1}
    \maketitle
    \begin{center}
	\vspace{-1em}
	\includegraphics[width=1.0\linewidth]{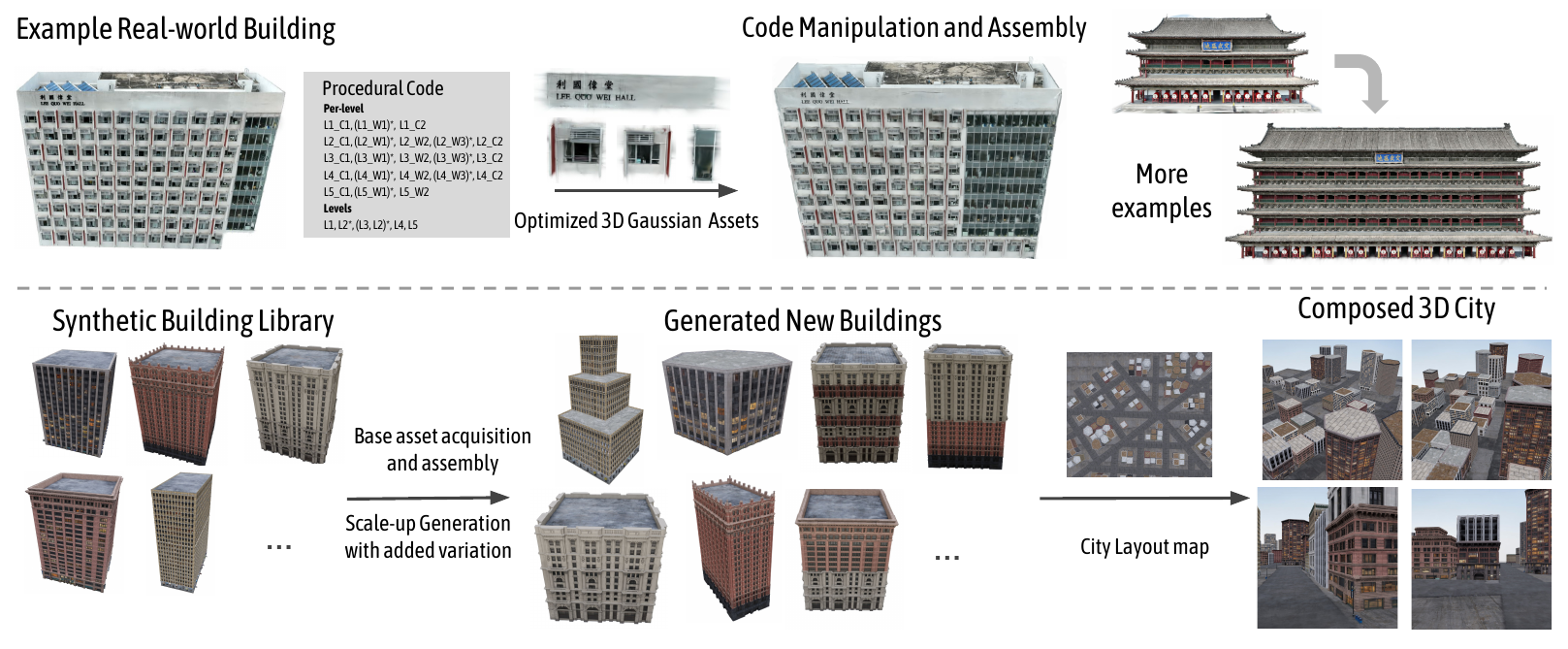}
    \vspace{-2em}
	\captionof{figure}{
    {Architectural structures in urban environments often exhibit repetitive patterns, such as the arrangement of windows and doors in buildings. Our approach focuses on extracting 3D base assets from predefined layouts and multi-view captures of buildings. Utilizing 3D-GS for reconstruction, we incorporate procedural code constraints during optimization to decompose the building entity into distinct 3D assets. 
    (1) The top row illustrates two real-world building examples, showcasing their appearances before and after editing. (2) The bottom row demonstrates the scaled-up city-level assembly achieved using UE's City Sample assets.
    ProcGS not only facilitates geometry editing but also enables the generation of new buildings by modifying or defining new procedural rules, offering a flexible, efficient, and scalable framework for city assembly with high-fidelity and precise control.}
    }
    % \vspace{-1em}
    \label{fig:teaser}
    \end{center}
}]
\begin{abstract}
Buildings are primary components of cities, often featuring repeated elements such as windows and doors. Traditional 3D building asset creation is labor-intensive and requires specialized skills to develop design rules. Recent generative models for building creation often overlook these patterns, leading to low visual fidelity and limited scalability. Drawing inspiration from procedural modeling techniques used in the gaming and visual effects industry, our method, Proc-GS, integrates procedural code into the 3D Gaussian Splatting (3D-GS) framework, leveraging their advantages in high-fidelity rendering and efficient asset management from both worlds. By manipulating procedural code, we can streamline this process and generate an infinite variety of buildings. This integration significantly reduces model size by utilizing shared foundational assets, enabling scalable generation with precise control over building assembly. We showcase the potential for expansive cityscape generation while maintaining high rendering fidelity and precise control on both real and synthetic cases. Project page:
\href{https://city-super.github.io/procgs/}{\textcolor{magenta}{https://city-super.github.io/procgs/}}.
\end{abstract}    
\section{Introduction}
\label{sec:intro}

High-quality 3D city assets are essential for virtual reality, video games, film production, and autonomous driving simulations. {While recent advances leverage deep generative models~\cite{lin2023infinicity, xie2024citydreamer, DBLP:journals/corr/abs-2406-06526, DBLP:journals/corr/abs-2407-11965,DBLP:journals/corr/abs-2312-01508} to scale up city generation process, the rendered visual fidelity and geometric accuracy of synthesized 3D city scenes remain unsatisfactory. Buildings, as pivotal elements in urban landscapes, pose a considerable challenge for generative models due to their intricate geometries and diverse appearances.} 
{Nevertheless, industries have a long history using procedural generation to create high-quality, diverse 3D structures, from architectural models to virtual cities, especially in film and game development. 
Typically, this procedure involves creating a set of base assets,~\eg window, corner and wall, designing procedural rules and configurations, and assemble assets accordingly. In such way, artists can construct scenes in a scalable fashion, as demonstrated in City Sample~\cite{CitySample} project.
However, creating base assets is non-trivial and requires considerable human efforts to create intricate meshes and textures.}

{Recently, 3D Gaussian Splatting (3D-GS)~\cite{kerbl20233d} has garnered significant attention for its photo-realistic visual quality, and efficient training and rendering. Its explicit nature make this representation interactable and becoming popular among various applications like VR/AR, game development, content creation etc. 
It is therefore attempting to adopt this representation in city generation tasks.
However, 3DGS has primarily been used to model entire scenes, which are usually compositions of multiple objects or elements, from multi-view captures. 
Isolating a specific part is challenging. For instance, while it’s straightforward to model an entire building from photos, focusing solely on each individual component of the building, becomes cumbersome and challenging.}

We present \emph{Proc-GS}, the first pipeline that enables procedural modeling with 3DGS. Our framework consists of two stages: (1) In the \emph{Asset Acquisition} stage, we constrain the optimization of 3DGS by guiding it to follow a predefined layout. For example, when modeling a building with 3DGS, we start by generating its procedural code, either manually or using an off-the-shelf segmentation model. This code is used to initialize a set of Gaussians for each base asset of the building. These asset-specific Gaussians are then assembled according to the procedural code, and we optimize the assembled Gaussians as a whole using rendering loss. Figure~\ref{fig:pipeline} illustrates an example. Notably, repeated base assets will be updated synchronously; to capture appearance various and subtle change in geometry, we additionally learn a variance code for each asset.
(2) In the \emph{Asset Assembly} stage, we use procedural code to manipulate base assets, generating buildings with diverse geometric structures and photorealistic appearances. We demonstrate that these newly created architectures can be integrated with Houdini~\cite{Houdini}, allowing for highly scalable scene composition with intuitive controls.

To showcase the capabilities of \emph{Proc-GS}, we curated the \emph{MatrixBuilding} dataset from the City Sample~\cite{CitySample}, which contains multi-view images and procedural codes for 17 iconic buildings. Our \emph{Proc-GS} approach enables flexible geometry editing and the creation of new structures by combining assets from different buildings, allowing users to generate vast, customized virtual cities. 
We also migrate \emph{Proc-GS} to real-world buildings, and enable the conversion from actual structures into virtual assets, supporting scalable, photo-realistic city generation that benefits games, autonomous driving, and embodied AI etc.
Experiments demonstrate that \emph{Proc-GS} outperforms previous city generation methods in both rendering and geometry quality.

In summary, our contributions are follows: 
 {
\begin{itemize}[left=0.2cm]
    \item \emph{MatrixBuilding Dataset}: A collection of dense multi-view images paired with procedural code for 17 iconic buildings, capturing high-resolution details and diverse architectural styles.
    \item \emph{Proc-GS}: The first framework that integrates procedural modeling with 3D-GS to accelerate 3D building asset creation, and extraction from the real world scenes. Our method enhances infinite city generation with high flexibility and photo-realistic visual quality.
\end{itemize}
}
\section{Related Work}
\subsection{Advancements in Neural Rendering}
 {
Neural rendering techniques, utilizing implicit representations for 3D modeling, have revolutionized novel view synthesis with photo-realistic rendering qualities. Recent advances fall into two broad categories: 1) differentiable volume rendering and 2) rasterization-based methods.
The most representative work of the former is NeRF~\cite{DBLP:conf/eccv/MildenhallSTBRN20}, which encodes the scene into the weights of Multi-Layer Perceptrons (MLPs). Despite its extraordinary ability to handle view-dependent appearance, the lack of explicit geometry structure hinders easy editing and physical interactions, making tasks like object insertion, deletion, and replacement tedious~\cite{tang2023delicate,haque2023instruct}. Additionally, the slow training and inference speed of NeRF and its variants~\cite{xu2022point,DBLP:conf/cvpr/BarronMVSH22} limits their practicality for large-scale generation. Even with more advanced backbones~\cite{DBLP:journals/tog/MullerESK22}, rendering efficiency and computational costs still lag behind traditional rasterization methods, limiting their real-time application potential.
In contrast, 3D Gaussian Splatting~\cite{kerbl20233d}, which projects 3D Gaussians to a 2D image plane via rasterization, achieves state-of-the-art rendering quality. The explicit nature of 3D Gaussians has enabled a variety of applications, including 3D generation~\cite{tang2023dreamgaussian}, physical simulation~\cite{xie2023physgaussian}, and editing~\cite{fang2023gaussianeditor}.} 

Inspired by the efficiency and high fidelity of 3D-GS, this work explores their potential for scalable generation of 3D scenes through procedural code and asset construction.

\subsection{Advances in City Generation}
{Recent advances in city generation have produced several notable approaches for creating complete urban environments. They could be categorized to three types: (1) whole city generation, (2) city layout generation and (3) city-view video generation. InfiniCity~\cite{lin2023infinicity} first introduced infinite-scale synthesis through a three-module system with octree-based voxel representation, followed by CityDreamer~\cite{xie2024citydreamer} which proposed a compositional approach separating building instances from background elements. CityGen~\cite{DBLP:journals/corr/abs-2312-01508} further advanced layout generation using outpainting and diffusion models, while recent methods like UrbanWorld~\cite{DBLP:journals/corr/abs-2407-11965} pioneered a comprehensive pipeline combining diffusion-based rendering with multimodal language models, and GaussianCity~\cite{DBLP:journals/corr/abs-2406-06526} adapted 3D Gaussian splatting with BEV-Point representation for efficient large-scale rendering. In addition, many methods contributes to the task of city layout generation by introducing multi-modal controllable generation~\cite{DBLP:journals/corr/abs-2407-17572}, a three-stage learning-based framework~\cite{DBLP:journals/corr/abs-2406-04983}, graph-based modeling~\cite{DBLP:journals/corr/abs-2407-11294} and a comprehensive dataset~\cite{DBLP:journals/corr/abs-2407-07835}. Finally, video diffusion models~\cite{DBLP:conf/siggraph/Deng0LGSW24} is also exploited for generating consistent city views.}

{Such generation-based methods commonly rely on generative priors to produce city views, yet their perceptual quality and 3D geometries could not be ensured. Our method pioneers the usage of procedural modeling and 3D-GS assets for better visual quality.
}
\subsection{Procedural Modeling as Scalable Generators}

 {
Procedural generation involves creating a vast variety of assets using generalized rules and simulators. This technique has garnered significant interest in the computer vision and graphics community due to its scalability and adaptability. It is extensively used for creating virtual environments~\cite{deitke2022️}, urban areas~\cite{chen2008interactive, merrell2023example, tsirikoglou2017procedural, parish2001procedural}, and natural landscapes~\cite{Raistrick_2023_CVPR, schott2023large, Khan_2019_CVPR_Workshops}. Additionally, procedural methods are employed for generating structured objects~\cite{nishida2018procedural, li2023rhizomorph, merrell2023example, makatura2023procedural} and textures~\cite{hu2023generating, greff2022kubric}.
Procedural generation functions as a powerful data simulator, particularly valuable when obtaining or generating high-quality real data is challenging. Traditional rule-based procedural generators are integrated into popular 3D modeling software such as Blender, Houdini, and Unreal Engine, thereby streamlining the creation workflow for artists. At its core, procedural modeling represents world-building through concise mathematical rules, allowing for complex and varied structures to be efficiently created and manipulated.
On the other hand, inverse procedural modeling addresses the challenge of inferring procedural rules from input data, either from 2D images~\cite{guo2020inverse, nishida2018procedural} or 3D models~\cite{demir2016proceduralization, mathias2011procedural}. This approach enables the extraction of procedural representations from existing assets, facilitating their integration into procedural workflows.
}

 {
Inspired by these advancements, we aim to introduce procedural properties into 3D Gaussian Splatting (3D-GS). By leveraging the advantages of procedural generation and 3D-GS, we can enhance the flexibility and scalability of 3D scene generation, enabling the creation of high-fidelity, expansive virtual environments.
}
\section{Procedural Modeling with 3D Gaussians}
\label{sec:method}

 {Our \emph{Proc-GS}  consists of two stages: (1) \emph{Asset Acquisition} uses procedural code during the training process of 3D Gaussian Splatting (3D-GS)~\cite{kerbl20233d} to acquire base assets; (2) \emph{Asset Assembly} manipulates the procedural codes to generate diverse buildings and assemble a 3D City. In the following sections, Sec.\ref{sec 3.1} introduces the basics of 3D-GS, Sec.\ref{sec 3.2} introduces the definition of procedural code, the synthetic \emph{MatrixBuilding} dataset and how to obtain procedural code from real world, Sec.~\ref{sec 3.3}  and Sec.~\ref{sec 3.4} separately introduces the \emph{Asset Acquisition} and \emph{Asset assembly}.
}

\subsection{Preliminaries: 3D Gaussian Splatting}\label{sec 3.1}
Unlike the implicit neural fields of NeRF~\cite{DBLP:conf/eccv/MildenhallSTBRN20}, 3D-GS~\cite{kerbl20233d} represents 3D scenes with explicit anisotropic 3D Gaussians, maintaining differentiability while enabling efficient tile-based rasterization.Initialized from a set of sparse point cloud, each 3D Gaussian $i$ is assigned with the learnable parameters $\{\mu_i, R_i, S_i, \alpha_i, \mathcal{C}_i\}$. For a given 3D point $x$ within the scene, 
% we have
\begin{equation}
G_i(x) = e^{-\frac{1}{2} (x-\mu_i)^T \Sigma_i^{-1} (x-\mu_i)}, \quad \Sigma_i = R_iS_iS_i^TR_i^T,
\label{gs}
\end{equation}
where $\mu_i$ represents the center of the Gaussian, $R_i$ and $S_i$ together define the covariance matrix $\Sigma_i$ of the Gaussian. 
The opacity is denoted by $\alpha_i$, which is multiplied by the Gaussian function $G_i(x)$ to determine the contribution of the final rendered output during the blending process. $\mathcal{C}_i$ represents the spherical harmonic coefficients used to obtain the view-dependent color. 

3D-GS uses tile-based rasterization for efficient scene rendering, projecting 3D Gaussians $G$ onto the image plane as 2D Gaussians $G'$. The rasterizer sorts these 2D Gaussians and applies $\alpha$-blending:
\begin{equation}
G'(x') = \sum_{i \in N} G_i(x') \alpha_i,
\label{alpha_blending}
\end{equation}
where $x'$ is the pixel position, $N$ is the number of 2D Gaussians, and $d_i$ is the view direction to the center of $G_i$. This differentiable rasterizer allows direct optimization of 3D Gaussian parameters under training view supervision. 3D-GS excels in creating high-quality 3D assets efficiently, with its discrete structure that simplifies editing and cross-scene transfer.
\begin{equation}
\small
C(x') = \sum_{i \in N} \text{SH}(\bm d_i; \mathcal{C}_i)  \sigma_i \prod_{j=1}^{i-1} (1 - \sigma_j), \quad \sigma_i = \alpha_i G'_{i}(x')
\label{gs_rendering}
\end{equation}

\begin{figure}[t]
\begin{center}
\includegraphics[width=1.0\linewidth]{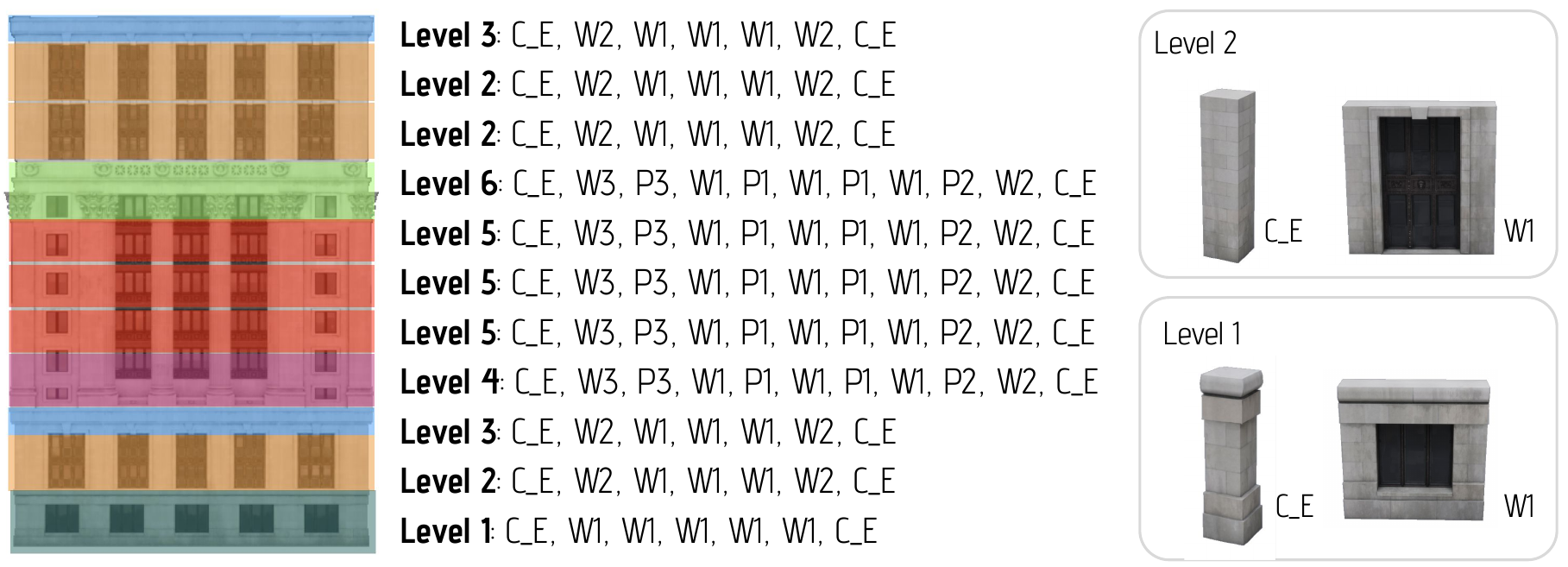}
\end{center}
\vspace{-1em}
\caption{\textbf{Example Building Procedural Code from City Sample~\cite{CitySample}.} Different levels are distinguished by colors, each represented by a string of characters indicating the instantiation of assets: \emph{C\_E} (external corner), \emph{P*} (pillar), and \emph{W*} (window). Base assets such as \emph{C\_E} and \emph{W1} for Levels 1 and 2, shown on the right, are manually created by artists.}
\label{fig:procecural_code}
\vspace{-1em}
\end{figure}

\subsection{Procedural Code Definition}
\label{sec:procedural_code_def}
\label{sec 3.2}
 Procedural modeling is widely used for building generation in game scenes, such as the City Sample in UE5~\cite{CitySample,UnrealEngine}, by leveraging buildings' structured nature and repetitive assets. Buildings are hierarchically decomposed into \emph{levels}, where identical layers may repeat, as shown by Levels 2, 3, and 5 in Figure~\ref{fig:procecural_code}. Each level consists of base assets like windows, corners, and pillars, with identical assets recurring within levels, such as \emph{C\_E} and \emph{W1} in Level 1. A building can thus be represented by a procedural code string and a set of base assets, where base assets are crafted manually by artists.  
 
 In this paper, we propose an alternative approach to extract base assets from multi-view images. To simplify the problem, we first assume the ground truth procedural code is provided. We created the \emph{MatrixBuilding} dataset based on 17 buildings from the City Sample~\cite{CitySample}, crafted by artists in Maya to emulate architectural styles from cities like New York, Chicago, and San Francisco. Each scene includes dense multi-view images, ground-truth camera poses, and ground-truth procedural codes. More details are provided in the Appendix~\ref{a.1}. Our framework can operate on synthetic worlds and is also practical for real-world scenarios. We discovered an efficient approach to obtain procedural code from real-world scenes. We first train 2D-GS~\cite{Huang2DGS2024} to obtain geometrically accurate point clouds and mesh. Subsequently, building facades are automatically estimated using the method proposed in ~\cite{Yu_2022_CVPR}. For each face, we render an image directly facing the building facade and annotate the procedural code on the 2D image. Then the 2D procedural code are projected onto the mesh to obtain the 3D procedural code. Please refer to the Appendix~\ref{b.1} for more details. Once the procedural code is obtained, our algorithm works almost identically for both real-world and synthetic scenes.

\subsection{Asset Acquisition}
\label{sec 3.3}
 {
In the gaming and animation industry, base assets are usually manually created by artists and assembled using either human-defined or heuristically generated procedural code. Our goal is to \emph{extract} these 3D base assets automatically during the training process of 3D-GS~\cite{kerbl20233d}. To achieve this, we assume procedural code is available, whether it's the ground truth code from the \emph{MatrixBuilding} dataset or estimated code from real-world scenes.
}

 {
In addition to the procedural code, as shown in Fig.~\ref{fig:procecural_code}, we obtain for each base asset: 
\begin{itemize}[left=0.2cm]
    \item the size of the asset's bounding box, $(x_e, y_e, z_e)$; 
    \item the pivot location, $(x_c, y_c, z_c)$ in its local coordinates;
    \item the set of transformations for $K$ instantiations in the world coordinate system $\{[R_1, T_1, S_1], [R_2, T_2, S_2], \ldots$, $ [R_K, T_K, S_K]\}$, where $T \in \mathbb{R}^{3\times1}$ is the translation vector, $R \in \mathbb{R}^{3\times3}$ is the rotation matrix and $S \in \mathbb{R}^{3\times1}$ is the scale factor.
\end{itemize} 
}

\medskip
\noindent{\bf Gaussian Initialization.} 
We initialize the pivot of each base asset at the origin of the world coordinate system, where the pivot is the origin of the asset's local coordinate system. The bounding box of the $i$-th base asset in the world coordinate system is represented as $(x^{i}_{min}, y^{i}_{min}, z^{i}_{min}, x^{i}_{max}, y^{i}_{max}, z^{i}_{max})$, where $x^{i}_{min} = x^{i}_{c} - \frac{x^{i}_{e}}{2}$ and $x^{i}_{max} = x^{i}_{c} + \frac{x^{i}_{e}}{2}$. The same calculation applies to the other two dimensions. The operations for synthetic and real-world scenes have subtle differences.

In synthetic scenes, for each building composed of a set of base assets $\mathcal{M}$, we initialize $N$ points. For the $i$-th asset, $N^i$ points are randomly initialized within its bounding box, determined by the ratio of the asset's bounding box volume to the total volume of all assets, as shown in Equation~\ref{equation:init}:
\begin{equation}
N^i = N * \frac{V^i}{\sum_{j\in\mathcal{M}} V^j}, \quad V^i=x^{i}_e \times y^{i}_e \times z^{i}_e.
\label{equation:init}
\end{equation}
No matter in real-world or virtual world, each instantiation of the $i$-th base asset will have minor differences in appearance and geometry, so we initialize a variance asset for each instantiation of the $i$-th base asset. For the $j$-th instantiation, we first randomly initialize $N^i$ points within the bounding box of $i$-th base asset. Then we update the center $\mu$, rotation $R$ and scale $S$ of the 3D Gaussians in this variance asset, as shown in Equation~\ref{equation:render}.

For real-world scenes, given the i-th base asset with $K$ instantiations, we apply transformations to map its bounding box to obtain the bounding boxes of all $K$ instantiations. For each instantiation, we filter SfM points within its bounding box for variance initialization. These point clouds are then transformed back to the world coordinate origin using inverse transformations of the i-th base asset and concatenated. Finally, we uniformly downsample the resulting point cloud by a factor of \emph{K} to initialize the i-th base asset.

\begin{figure*}[t]
\begin{center}
\includegraphics[width=1.0\linewidth]{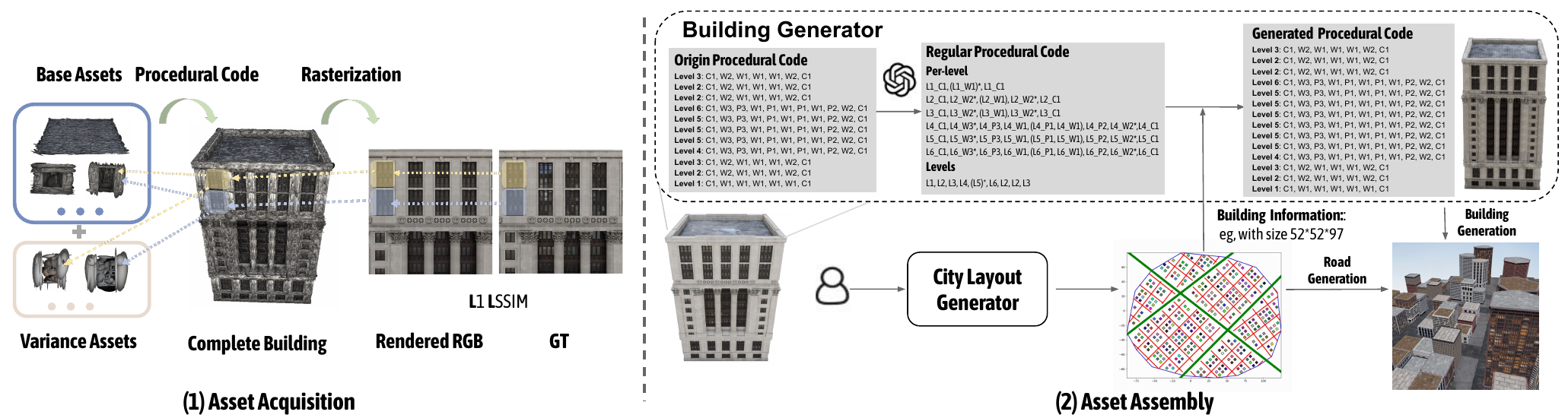}
\end{center}
\vspace{-1em}
\caption{{\bf Overview of ProcGS.} 
 { Our pipeline consists of two stages: (1) \textbf{Asset Acquisition}: We acquire the base assets in the training process of the 3D-GS. These assets are then assembled according to procedural code and add variance assets to create a complete building, which is used for novel view synthesis with Gaussian Splatting.
 (2) \textbf{Asset Assembly}: We use the \emph{building generator} and \emph{city layout generator} to assemble these base assets into a vivid 3D city.
 Users provide basic urban spatial data (purple city boundary and green primary road) to city layout generator to automatically predict other roads and building layouts. The building generator then assembles base assets into complete buildings using procedural code and predicted building parameters.}
}
\label{fig:pipeline}
\vspace{-1em}
\end{figure*}

\medskip
\noindent{\bf Rendering with Procedural Code.} Figure~\ref{fig:pipeline} (1) illustrates our rendering pipeline. First, we assemble the base assets according to the procedural code. The $i$-th base asset has a set of 3D Gaussians $\mathcal{A}_i$ and transformations $\{[R^{i}_{1}, T^{i}_{1}, S^{i}_{1}],$ $[R^{i}_{2}, T^{i}_{2}, S^{i}_{2}],$ $\ldots,$ $[R^{i}_{j}, T^{i}_{j}, S^{i}_{j}]$, $\ldots$,$ [R^{i}_{K}, T^{i}_{K}, S^{i}_{K}]\}$. For the $j$-th instantiation of the $i$-th base asset, the 3D Gaussian properties remain the same except for the center $\mu$, rotation $R$ and scale $S$, which are updated as:
\begin{equation}
\mu' = R^{i}_{j} \cdot S^{i}_{j}\cdot \mu + T^{i}_{j}, \quad R' = R^{i}_{j} \cdot R, \quad S' = S^{i}_{j} \cdot S.
\label{equation:render}
\end{equation}
After instantiating all the base assets according to the procedural code and adding the variance assets, the complete building is formed, as shown in Figure~\ref{fig:pipeline} (1).
The 3D Gaussians of the complete building are then fed into the rasterizer to render the image, supervised by the training views. A base asset may appear multiple times within the same image. During optimization, gradients from these repeated instances are backpropagated to the shared base asset and respective variance assets, refining the Gaussian parameters.

\begin{figure*}[t]
\begin{center}
\includegraphics[width=0.8\linewidth]{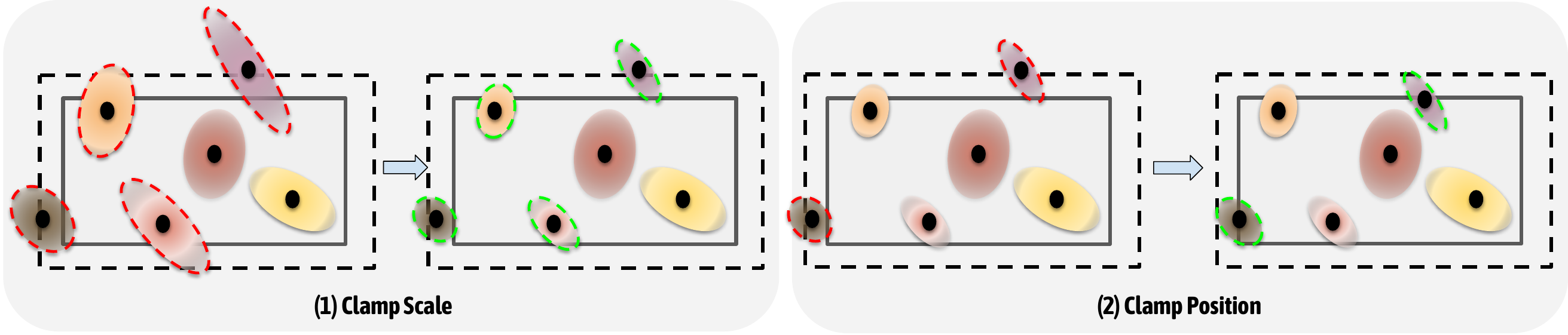}
\end{center}
\vspace{-1em}
\caption{{\bf Bbox Adaptive Clamp}. 
 {
(1) Clamping Scale: if a Gaussian exceeds the soft bounding box, the scale of this Gaussian is reduced by half.
(2) Clamping Position: repositions Gaussians that exceed the bounding box, realigning them within the bounding box.}
}
\label{fig:prune_bbox}
\end{figure*}

\noindent{\bf Bbox Adaptive Control of Gaussians.}
The internal ordering of Gaussians can be chaotic, often leading to good rendering results but disordered boundaries for base assets, complicating subsequent editing and generation (Figure~\ref{fig:prune_effect}). To address this, we enhance the original 3D-GS~\cite{kerbl20233d} with a \emph{Bbox Adaptive Clamp} operation (Figure~\ref{fig:prune_bbox}) in addition to densification and pruning. 
For each base asset, the \emph{Bbox Adaptive Clamp} operation involves:
(1) Clamp Scale: Using a slightly larger `soft' bounding box to avoid over-clamping. If a Gaussian exceeds the soft box boundaries, its scale is halved.
(2) Clamp Position: Pulling the centers of Gaussians exceeding the bounding box back to its edge.
This operation is performed every 100 iterations for both base and variance assets to maintain ordered boundaries and facilitate efficient extraction and manipulation.

\noindent{\bf Loss Functions.} 
We optimize the learnable Gaussians' parameters with respect to the $\mathcal{L}_1$ loss over rendered pixel colors and SSIM term~\cite{DBLP:journals/tip/WangBSS04} $\mathcal{L}_{\text{SSIM}}$. The
total supervision is given by
$\mathcal{L} =\mathcal{L}_1 + \lambda_{\text{SSIM}}\mathcal{L}_{\text{SSIM}}$.

\subsection{Asset Assembly}
\label{sec 3.4}

After extracting base assets from multi-view images, we can manipulate procedural code to generate new buildings. Combined with a rule-based 2D city layout generator, we can assemble complete 3D city scenes. As shown in Figure~\ref{fig:pipeline} (2), the assembly process consists of a building generator and a city layout generator.

\noindent{\bf Building Generator} For architectural generation, we first analyze the arrangement patterns of base assets within each floor and between floors in the original building, converting them into regular procedural code through GPT-4o~\cite{hurst2024gpt} with several examples provided in the prompt. The detailed process is explained in the Appendix~\ref{c.1}. In the regular procedural code, repeatable and scalable combinations both within and between floors are specified. Concretely, combinations within $\left( \quad \right)$ are designed as repeatable elements, while assets marked with $*$ are scalable to fit the size of the building. Subsequently, we can place assets according to the procedural code and specified building size (e.g., length, width and height) to generate new buildings with varying arrangements and sizes. Notably, during the building generation process, each base asset is randomly assigned a corresponding variance asset to enhance diversity and realism. Besides, we could also create new buildings from base assets extracted from different architectural sources. 

\noindent{\bf City Layout Generator} For city generation, users first select boundary points of the city map (purple boundary) and primary roads' endpoints (green lines), as shown in Figure~\ref{fig:pipeline} (2). Subsequently, we partition the city into several connected blocks based on the primary roads, with each block being randomly assigned regional characteristics (e.g., distribution of building sizes). Following this, we generate perpendicular secondary roads and determine building positions, topological structures, and sizes according to predefined rules. Finally, we randomly place decorative elements along the roads, such as street lamps, garbage bins, and mailboxes, which are also collected by the 3D-GS.

\section{Experiments}
\subsection{Experimental Setup}
\label{sec:dataset_collect}
\noindent{\bf Dataset.}  We collected the \emph{MatrixBuilding} dataset, featuring dense multi-view images of 17 buildings and their corresponding ground truth procedural codes from the City Sample~\cite{CitySample}.
Following MatrixCity~\cite{DBLP:conf/iccv/0002JXX0L023}, we disabled motion blur and used anti-aliasing during rendering to achieve the highest possible image quality. Details for each building and camera capture trajectory are provided in the Appendix~\ref{a.1}. We also validated our method on three real-world scenes captured by drones.

\noindent{\bf Implementations.}
We selected 3D-GS~\cite{kerbl20233d} as our primary baseline due to its state-of-the-art performance in novel view synthesis. Both 3D-GS and our proposed method were trained for 30k iterations and densified until 15k iterations. The number of gaussians initialized for synthetic data, $N$, in our method is set to 10k, as illustrated in Equation~\ref{equation:init}. 3D-GS is initialized with the building assembled by the randomly initialized base assets according to the procedural codes. The soft bounding box is expanded by 20cm beyond the bounding box. The loss weight $\lambda_{\text{SSIM}}$ is set to 0.2. Our model is trained on single RTX 3090 GPU.

\noindent{\bf Metrics.}  For novel view synthesis, we report widely adopted metrics: PSNR, SSIM~\cite{DBLP:journals/tip/WangBSS04}, and LPIPS~\cite{DBLP:conf/cvpr/ZhangIESW18}. Additionally, we report the number of Gaussians to evaluate model compactness. The metrics are averaged across all scenes for quantitative comparison. For city generation, we report {\bf Depth Error} (DE) and {\bf Camera Error} (CE) following EG3D~\cite{DBLP:conf/cvpr/ChanLCNPMGGTKKW22} and CityDreamer~\cite{xie2024citydreamer} to evaluate the 3D scene geometry and consistency. For DE, we utilize a pre-trained model~\cite{DBLP:journals/pami/RanftlLHSK22} to estimate the depth maps from the rendered images and calculate the $\ell_2$ distance between the normalized estimated depth and rendered depth. For CE, we first render images using hemispherically sampled camera poses, then estimate these poses using COLMAP~\cite{DBLP:conf/cvpr/SchonbergerF16}. The camera error is computed as a scale-invariant $\ell_2$ loss between the estimated and ground truth poses. 

\begin{table}[t]

\centering
\resizebox{0.99\linewidth}{!}{
\begin{tabular}{c|c|ccccc}
\toprule
Data Type & Method & PSNR $\uparrow$ & SSIM $\uparrow$ & LPIPS $\downarrow$ & \#GS (k) $\downarrow$\\ \midrule
\multirow{2}{*}{Synthetic} &{3D-GS} & 27.54 & {0.910} & {0.108} & {1,238} \\
&{Proc-GS (ours)} & \textbf{{27.68}} & \textbf{0.917} & \textbf{0.102} & \textbf{291} \\
\midrule
\multirow{2}{*}{Real} &{3D-GS} &\textbf{27.38} & \textbf{0.858}&\textbf{0.192}& 500 \\
&{Proc-GS (ours)} &27.19&0.853&0.196&\textbf{384} \\
\bottomrule
\end{tabular}}
\caption{\textbf{Quantitative comparisons.} {Proc-GS achieves similar perceptual quality while requiring less memory and providing flexible control capabilities, demonstrating the effectiveness of integrating procedural code with 3D-GS for building scenes.}}
\label{tab:comparison}
\end{table}

\begin{figure*}[h]
\begin{center}
\includegraphics[width=1.0\linewidth]{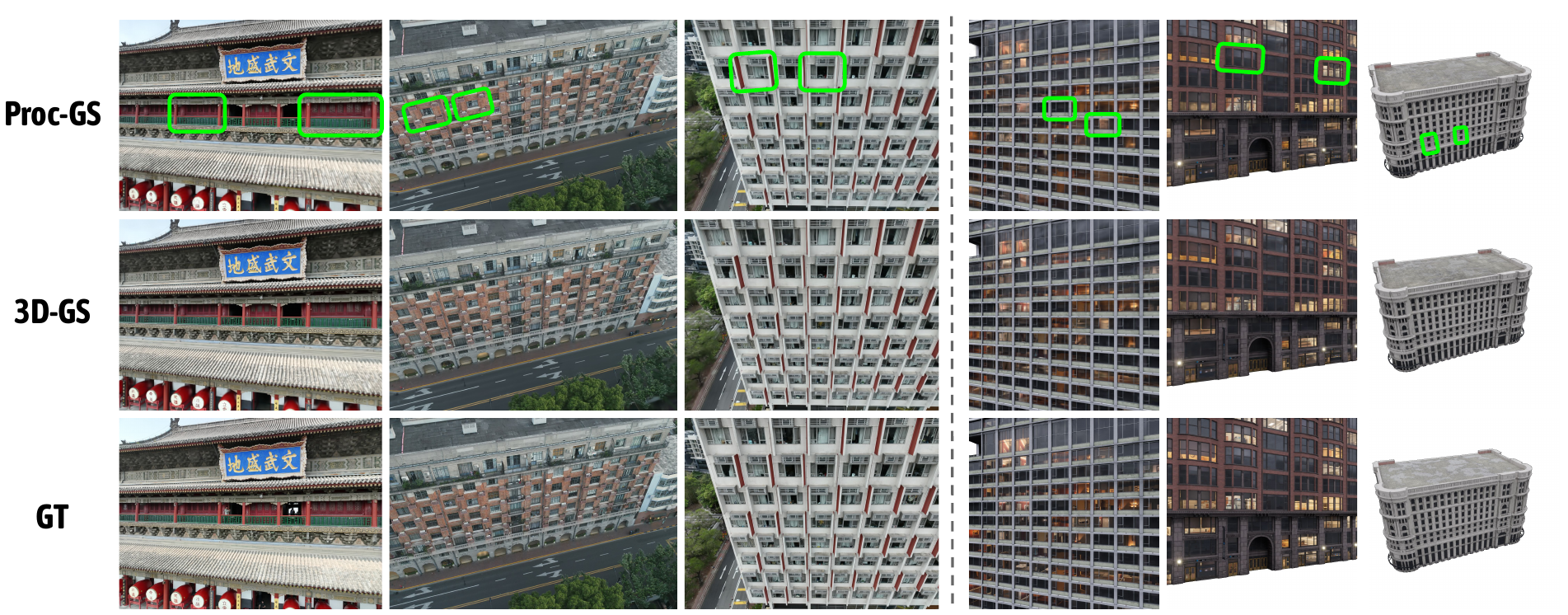}
\end{center}
\vspace{-1em}
\caption{\textbf{Qualitative results.} The left section shows results from three real-world scenes, while the right section presents results from the \emph{MatrixBuilding} dataset. Proc-GS achieves rendering quality comparable to 3D-GS. Green boxes in each image highlight pairs of instantiations that share the same base assets, illustrating our method's capability to effectively model variations in geometry and appearance.}
% \vspace{-2.em}
\label{fig:qualitative}
% \vspace{-2pt}
\end{figure*}

\iffalse
\begin{table*}[t]
\caption{\textbf{Sparse view quantitative results.} Proc-GS is robust to the view elimination and outperforms the baseline by a large margin under sparse-view setting, demonstrating the fitness to optimization.}
\centering
% \vspace{0.5em}
\setlength{\tabcolsep}{20pt}
\resizebox{\linewidth}{!}{
\begin{tabular}{c|ccc|ccc}
\toprule
  & \multicolumn{3}{c|}{\textbf{3D-GS} ~\cite{kerbl20233d}}                                                                & \multicolumn{3}{c}{{\textbf{Proc-GS (ours)}}}                                     \\
\multirow{-2}{*}{Trainning Views}   & PSNR  $\uparrow$                       & SSIM  $\uparrow$                       & LPIPS  $\downarrow$                      & PSNR $\uparrow$                        & SSIM  $\uparrow$                       & LPIPS $\downarrow$                       \\ \midrule
12 & {14.14} & {0.5120} & {0.4465} & {23.21} & {0.8175} & {0.1945} \\
24 & {19.41} & {0.6868} & {0.3131} & {26.94} & {0.8906} & {0.1374} \\
47                  & {27.09} & {0.8680} & {0.1688} & {28.92} & {0.9238} & {0.1145} \\
469                  & {33.60} & {0.9561} & {0.0930} & {29.73} & {0.9414} & {0.1013} \\
\bottomrule
\end{tabular}}
\label{tab:sparse}
% \vspace{-1.5em}
\end{table*}
\fi

\begin{table}[t]
\centering
% \vspace{0.5em}
\setlength{\tabcolsep}{8pt}
\resizebox{\linewidth}{!}{
\begin{tabular}{c|ccc|ccc}
\toprule
 & \multicolumn{3}{c|}{\textbf{3D-GS} ~\cite{kerbl20233d}} & \multicolumn{3}{c}{\textbf{Proc-GS}} \\
\multirow{-2}{*}{Views} & PSNR  $\uparrow$ & SSIM  $\uparrow$ & LPIPS  $\downarrow$ & PSNR $\uparrow$ & SSIM $\uparrow$ & LPIPS $\downarrow$ \\ 
\midrule
24 & {16.93} & {0.542} & {0.410} & {\bf 19.70} & {\bf 0.682} & {\bf 0.294} \\
47                  & {20.65} & {0.688} & {0.283} & {\bf 23.11} & {\bf 0.795} & {\bf 0.196} \\
469                  & {27.54} & {0.910} & {0.108} & {\bf 27.68} & {\bf 0.917} & {\bf 0.102}  \\
\bottomrule
\end{tabular}}
\caption{\textbf{Sparse view quantitative results.} Proc-GS is robust to the view elimination and outperforms the baseline by a large margin under sparse-view setting on \emph{MatrixBuilding} Dataset, showing the fitness to optimization.}
\label{tab:sparse}
\vspace{-1.em}
\end{table}

\subsection{Comparisons with 3D-GS}
In Table~\ref{tab:comparison} and Figure~\ref{fig:qualitative}, we compare our Proc-GS with the original 3D-GS on both synthetic and real-world scenes. For synthetic data, Proc-GS achieves comparable novel view synthesis quality while significantly reducing the model size by a factor of 4, demonstrating the effectiveness of our approach. For real-world scenes, due to their complex appearance, geometry, and lack of ground-truth procedural code, the optimization becomes more challenging and requires more Gaussians for the variance assets. Although the compression rate is lower than in synthetic cases, our model still maintains a smaller size compared to 3D-GS. More importantly, we gain flexible control capability with only a slight decrease in accuracy, which is barely noticeable in the qualitative results (Figure~\ref{fig:qualitative}).

Another significant strength of our Proc-GS is its ability to reconstruct from sparse view inputs, as depicted in Table~\ref{tab:sparse}. Using the dense view benchmark of 3D-GS, our Proc-GS delivers similar results with 3D-GS on \emph{MatrixBuilding} dataset. However, 3D-GS with fewer views delivers significantly inferior results compared to the dense view scenario. In contrast, our Proc-GS maintains robustness against the reduction in the number of training views since the repeated occurrence of base assets naturally serves as a form of data augmentation. More qualitative comparisons are provided in the Appendix~\ref{d.1}. The robustness of Proc-GS with sparse views shows great potential for real-world applications as the data collection process is cumbersome and there will be many scenes that very limited views of data is accessible.

\subsection{Ablation Studies}

\begin{figure}[t]
\begin{center}
\includegraphics[width=0.97\linewidth]{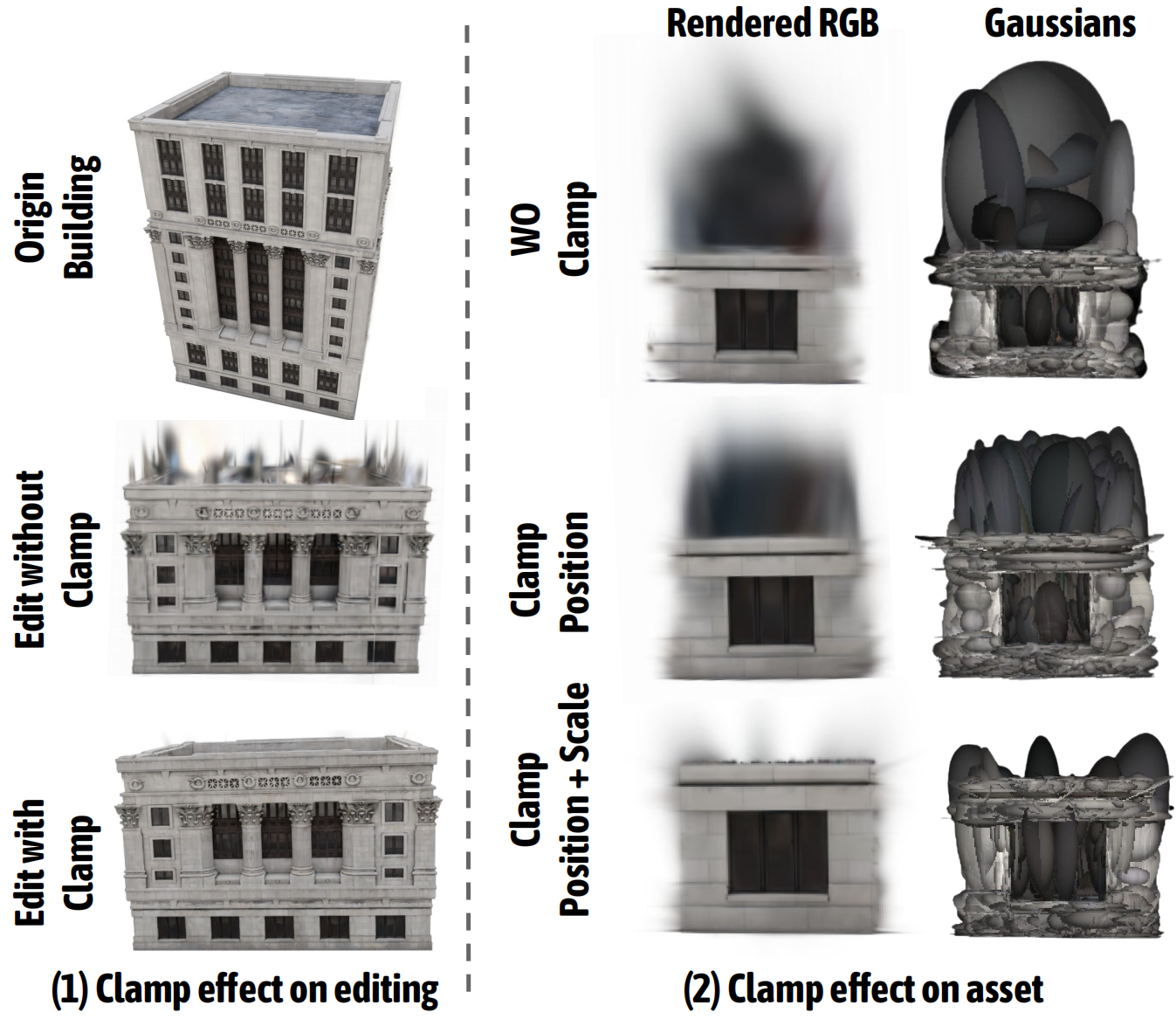}
\end{center}
\vspace{-1em}
\caption{\textbf{(1) Clamp effect on editing}. Without clamp, the boundary area of edited scene is intensively corrupted by artifacts, making it impractical to create a new building with these assets; \textbf{(2) Clamp effect on asset}. We ablate effects of the clamp operation and demonstrate the effectiveness of both strategies.}
\vspace{-1em}
% \vspace{-10pt}
\label{fig:prune_effect}
% \vspace{-2pt}
\end{figure}

\begin{figure*}[ht]
\begin{center}
\includegraphics[width=1.0\linewidth]{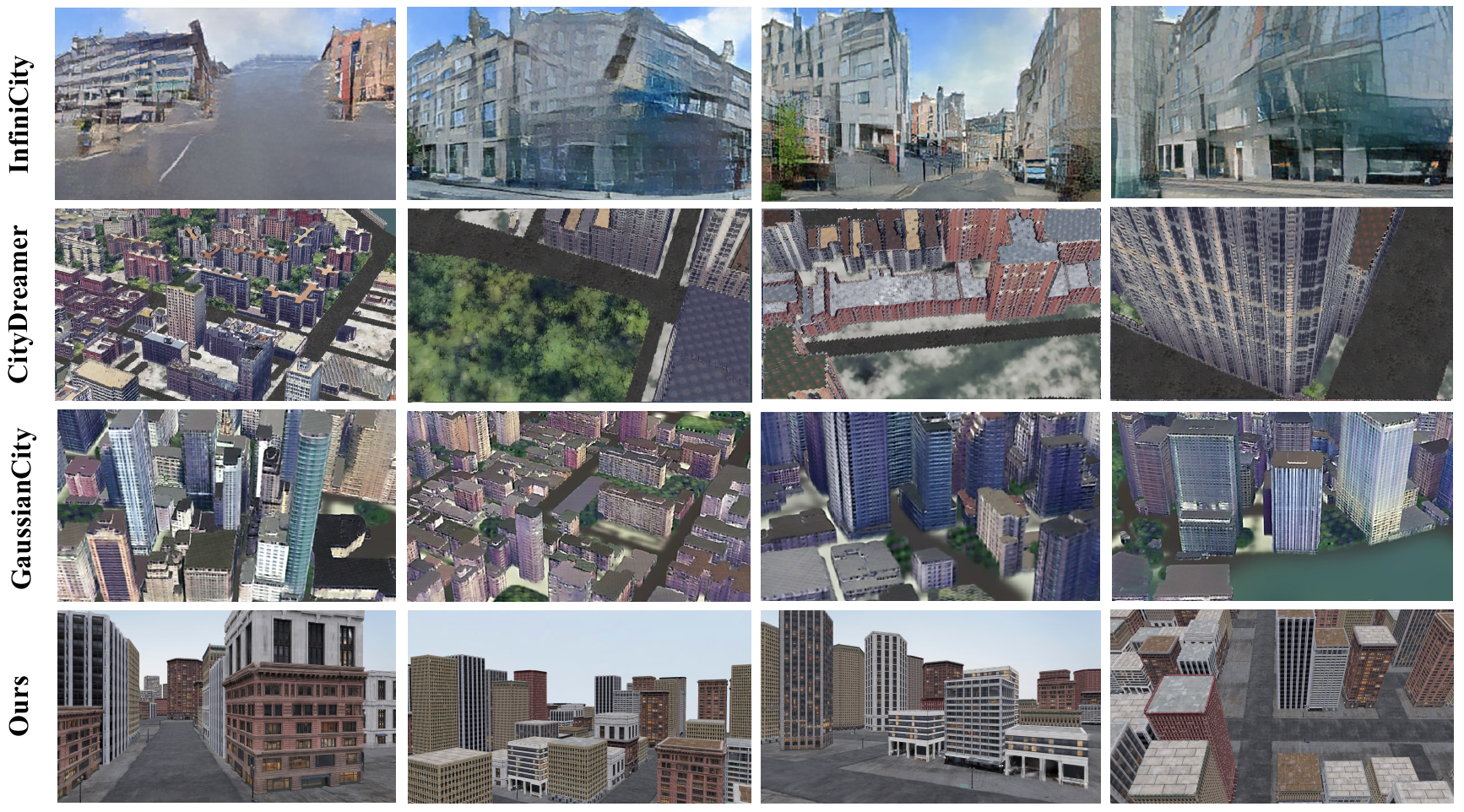}
\end{center}
\vspace{-1em}
\caption{
{
\textbf{Qualitative comparisons of city generation results}. Our Proc-GS framework learns base assets during the training process of 3D-GS using procedural codes, which are then manipulated to assemble these assets into a cohesive 3D city. Compared to other generation-based methods, Proc-GS demonstrates superior visual quality in both aerial and street-level views, especially in architectural details. We recommend zoom-in for detailed inspection.
}
}
\label{fig:city}
% \vspace{-2pt}
\end{figure*}

\begin{table}[h]

\centering
\resizebox{\linewidth}{!}{
\begin{tabular}{c|ccc|cccc}
\toprule
 & PC & BC & VR & PSNR$\uparrow$ & SSIM  $\uparrow$ & LPIPS  $\downarrow$ &  \#GS (k) $\downarrow$ \\ \midrule
1 & &  &  &  {27.54} & {0.910} & {0.108} & {1238}\\
2 & $\checkmark$ & & &25.54 &  0.903 &  0.123 & 87\\ 
3 &$\checkmark$ & $\checkmark$ &  &24.40  & 0.892  & 0.132 & 87\\ 
4 &$\checkmark$ & $\checkmark$ & $\checkmark$ & \textbf{27.68}  & \textbf{0.917}  & \textbf{0.102} & \textbf{291}\\ 
\bottomrule
\end{tabular}}
% \vspace{-1em}
\caption{\textbf{Quantitative ablations} on Procedural Code (PC), Box Adaptive Clamp (BC), and Variance (VR) using the \emph{MatrixBuilding} dataset. Row 1 indicates the vanilla 3D-GS baseline.}
\vspace{-1.5em}
\label{tab:ablation}
\end{table}

\noindent{\textbf{Clamp strategies.} Building components are seamlessly connected. But during component decoupling, Gaussian kernels often extend beyond asset boundaries. This overlap complicates asset combination and building editing, as demonstrated in Figure~\ref{fig:prune_effect} (1). To address this issue, we developed clamp strategies as illustrated in Figure~\ref{fig:prune_bbox}. Figure~\ref{fig:prune_effect} (2) shows a qualitative evaluation of our clamp operations in Proc-GS. The implementation of these two clamp strategies results in significantly cleaner asset boundaries.}

\noindent{\textbf{Variance assets.} However, as shown in Table~\ref{tab:ablation}, we find that integrating with the procedural code and adding the clamp operations both leads to worse rendering quality. This is because the instantiations of base assets are not completely identical. When different instantiations do not affect each other at all, the ability to model diversity is minimized. Therefore, we have added a variance asset to each instantiation, which allows us to achieve similar performance as 3D-GS. In the \emph{Asset Assembly} process, we randomly assign variance assets to each instantiation to enhance the diversity of generated buildings.}

\iffalse
\begin{table}[t]
\caption{Sparse view results.}
\centering
% \vspace{-1em}
\resizebox{0.7\linewidth}{!}{
\begin{tabular}{c|cccc}
\toprule
   & PSNR  $\uparrow$                       & SSIM  $\uparrow$                       & LPIPS  $\downarrow$      \\ \midrule
{Without Clamp} & {32.55} & {0.9533} & {0.0922} \\
{Clamp Position}& {31.64} & {0.9478} & {0.1003} \\
{Clamp Position and Scale (ours)} & {29.73} & {0.9414} & {0.1013} \\
\bottomrule
\end{tabular}}
% \vspace{-1em}
\label{tab:clamp}
\end{table}
\fi

\subsection{Results on City Generation}
\begin{table}
\centering
\resizebox{0.8\linewidth}{!}{
\begin{tabular}{lcc}
\toprule
Method & Camera Error$\downarrow$ & Depth Error$\downarrow$ \\
\midrule
Pers. Nature~\cite{DBLP:conf/cvpr/Chai0LIS23} & 86.371 & 0.109 \\
SceneDreamer~\cite{DBLP:journals/pami/ChenWL23} & 0.186 & 0.216 \\
CityDreamer~\cite{xie2024citydreamer} & 0.060 & 0.096 \\
GaussianCity~\cite{DBLP:journals/corr/abs-2406-06526} & 0.057 & 0.090 \\
Ours & {\bf 0.049} & {\bf 0.032} \\
\bottomrule
\end{tabular}
}
\caption{{\bf Quantitative comparison} of Depth and Camera Error. Our method outperforms existing approaches.}
\vspace{-1.5em}
\label{tab:depth_camera}
\end{table}

Our Proc-GS can extract base assets in the training process of 3D-GS based on procedural codes. By manipulating procedural codes, we can flexibly edit the building geometry, generate new building using cross-scene base assets and assemble these buildings into city scenes as shown in Figure~\ref{fig:teaser}.  More qualitative results are provided in the Appendix~\ref{d.2}. In Table~\ref{tab:depth_camera}, we compare our assembled city with other baseline methods quantitatively. Our approach achieves better scores on CE and DE metrics, indicating superior 3D consistency in the generated city scenes. We compare with other methods qualitatively in Figure~\ref{fig:city}. We demonstrate enhanced visual quality, more stable generation, and greater flexibility in control.

\section{Limitations and Potentials}
\label{sec:limitation}
While we have demonstrated the possibility to obtain high-quality base assets from real scenes, scaling up this process faces several challenges: 1) fully automating procedural code generation without any human intervention; 2) extracting high-quality base assets from sparse views or even single images to reduce data collection costs. Additionally, our current layout generation does not consider aesthetic principles or urban functionality. In the future, LLMs could replace rule-based systems to generate more practical and aesthetically pleasing urban layouts.
{This will allow us to collect a wealth of base assets from real scenes and automatically assemble them into virtual cities, enhancing practical applications that rely on photorealistic data, such as embodied AI and autonomous driving.}
\section{Conclusion}

In this paper, we propose Proc-GS, a novel approach to efficiently craft high-quality base building assets by utilizing procedural code during the 3D Gaussian Splatting (3D-GS) training process. Proc-GS decomposes a complete Gaussian model of a building into base assets and a procedural code string. This decomposition enables flexible editing of building geometries and the creation of diverse structures by combining base assets from different scenes, thereby supporting the generation of extensive cityscapes. Proc-GS leverages the efficient rendering capabilities and the discrete structure of 3D-GS, and demonstrates its versatility on both synthetic and real-world scenarios. 
{
    \small
    \bibliographystyle{ieeenat_fullname}
    \bibliography{main}
}
\clearpage
% \setcounter{page}{1}
% \maketitlesupplementary
\maketitleappendix
\appendix

% \section{Appendix / supplemental material}

\section{More Dataset Details}\label{a.1}
\label{sec:supp_dataset}
\begin{figure*}[t]
\begin{center}
\includegraphics[width=1.0\linewidth]{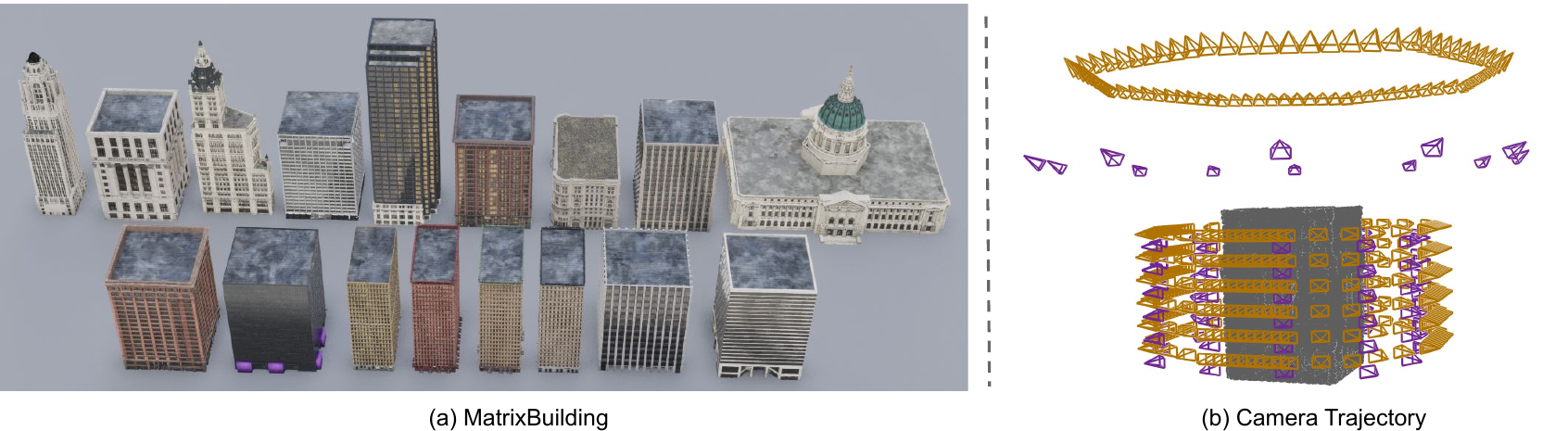}
\end{center}
\caption{\textbf{Dataset Overview.} (a) Overview of the 17 buildings in our proposed \emph{MatrixBuilding} dataset (b) Yellow cameras represent training views and purple cameras represent test views. The proportion of training views to test views is about 5:1.}
\label{fig:dataset_overview}
% \vspace{-2pt}
\end{figure*}

Our $MatrixBuilding$ Dataset consist of 17 buildings from the City Sample Project~\cite{UnrealEngine,CitySample}, as shown in Figure~\ref{fig:dataset_overview} (a), which contains ground-truth procedural code and dense multi-view images. These buildings are created to mimic the building styles of Chicago, San Francisco, and New York. In Table~\ref{tab:statistic}, we also provide the number of building base assets and the total count of instantiated assets after assembling complete buildings according to procedural codes. The design of base assets combined with procedural codes significantly reduces the model size. Figure~\ref{fig:dataset_overview} (b) shows the dense camera capture trajectories. The ratio between training view and test view is about five to one.

\begin{table*}[t]
\centering
% \vspace{0.5em}
\resizebox{\linewidth}{!}{
\begin{tabular}{c|ccccccccccccccccc}
\toprule
   Building & CHB                       & CHD                       & CHE  &CHF                      & CHG & CHH& CHI                       & CHJ                       & NYAA  &NYAB                      & NYAE & NYAF& NYG                       & SFA                       & SFB  &SFD                      & SFE\\ \midrule
\# Base Assets &90 & 30& 90& 32& 24& 19&43 & 8& 17& 24& 25& 37&56 & 54& 81& 12& 20\\
\# Total Assets & 1559& 345& 1645& 1170& 617& 1585& 697& 2409& 1869& 1920& 1929& 2831& 438& 295& 821& 729& 1405\\
\bottomrule
\end{tabular}}
% \vspace{-0.5em}
\caption{\textbf{Base Assets Statistics.} $CH*$ means a building of Chicago. $SF*$ means a building of San Francisco. $NY*$ means a building of New York. \# Total Assets means the total count of instantiated assets after assembling complete buildings according to the procedural codes }
\label{tab:statistic}
\end{table*}

\section{Exploration of Real-World Scenes}\label{b.1}
\begin{figure*}[t]
\begin{center}
\includegraphics[width=1.0\linewidth]{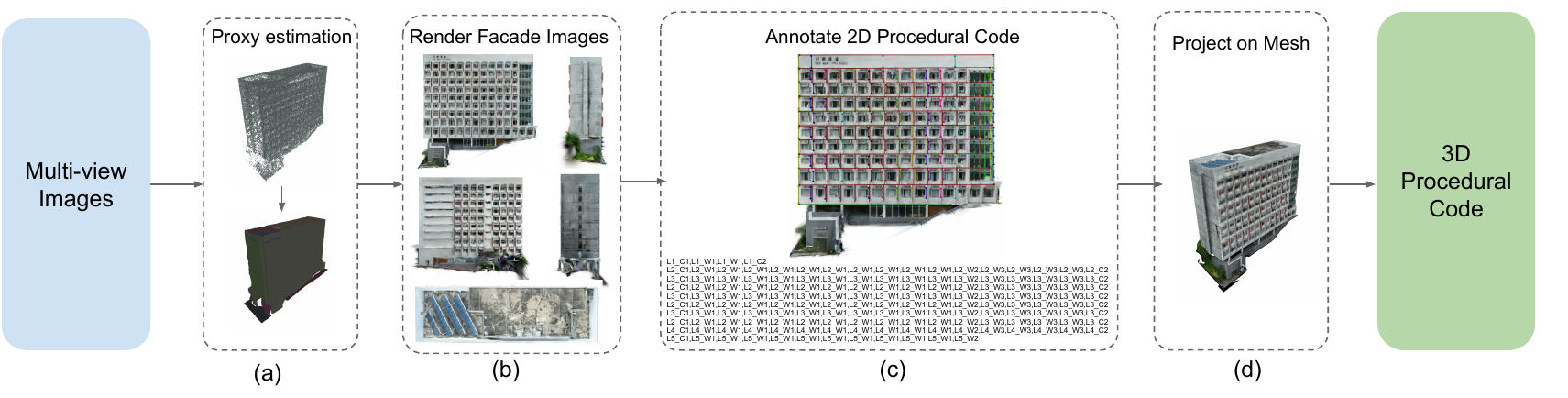}
\end{center}
\caption{\textbf{Extracting Procedural Code from Real-World Scenes.} (a) We extract point clouds with good geometric structures from multi-view images using 2D-GS~\cite{Huang2DGS2024}.Then we use the method~\cite{Yu_2022_CVPR} to automatically estimate the building facade. (b) For each facade, automatically render a maximally large image with comprehensive information directly facing the facade. (c) Manually annotate 2D procedural code for each facade. (d) Project the 2D procedural code onto the 2D-GS mesh to derive the corresponding 3D procedural code.}
\label{fig:real_data}
% \vspace{-2pt}
\end{figure*}

Figure~\ref{fig:real_data} illustrates our approach to extracting procedural code from real-world scenes. We begin by utilizing 2D-GS~\cite{Huang2DGS2024} to extract point clouds with good geometric structures from multi-view images, followed by automatically estimating building facades using the method of Yu et al.~\cite{Yu_2022_CVPR}. Buildings are composed of multiple facades. For each facade, we automatically render a maximally large image with comprehensive information directly facing the facade. Subsequently, we manually annotate 2D procedural code for each facade, and then project these annotations onto the 2D-GS mesh to derive the corresponding 3D procedural code. For each instantiation of the base assets, the range of the bounding box in the z-direction is calculated based on the current facade position and empirically set facade thickness. Automatically obtaining 2D procedural code could potentially be replaced by segmentation methods, which enables acquiring base assets from real-world scenes with minimal human intervention, beyond the initial data collection effort. We will conduct in-depth exploration of this direction in the future to enable large-scale collection of base assets from real-world scenes.

\section{Prompt Example}\label{c.1}
\begin{figure*}[t]
\begin{center}
\includegraphics[width=1.\linewidth]{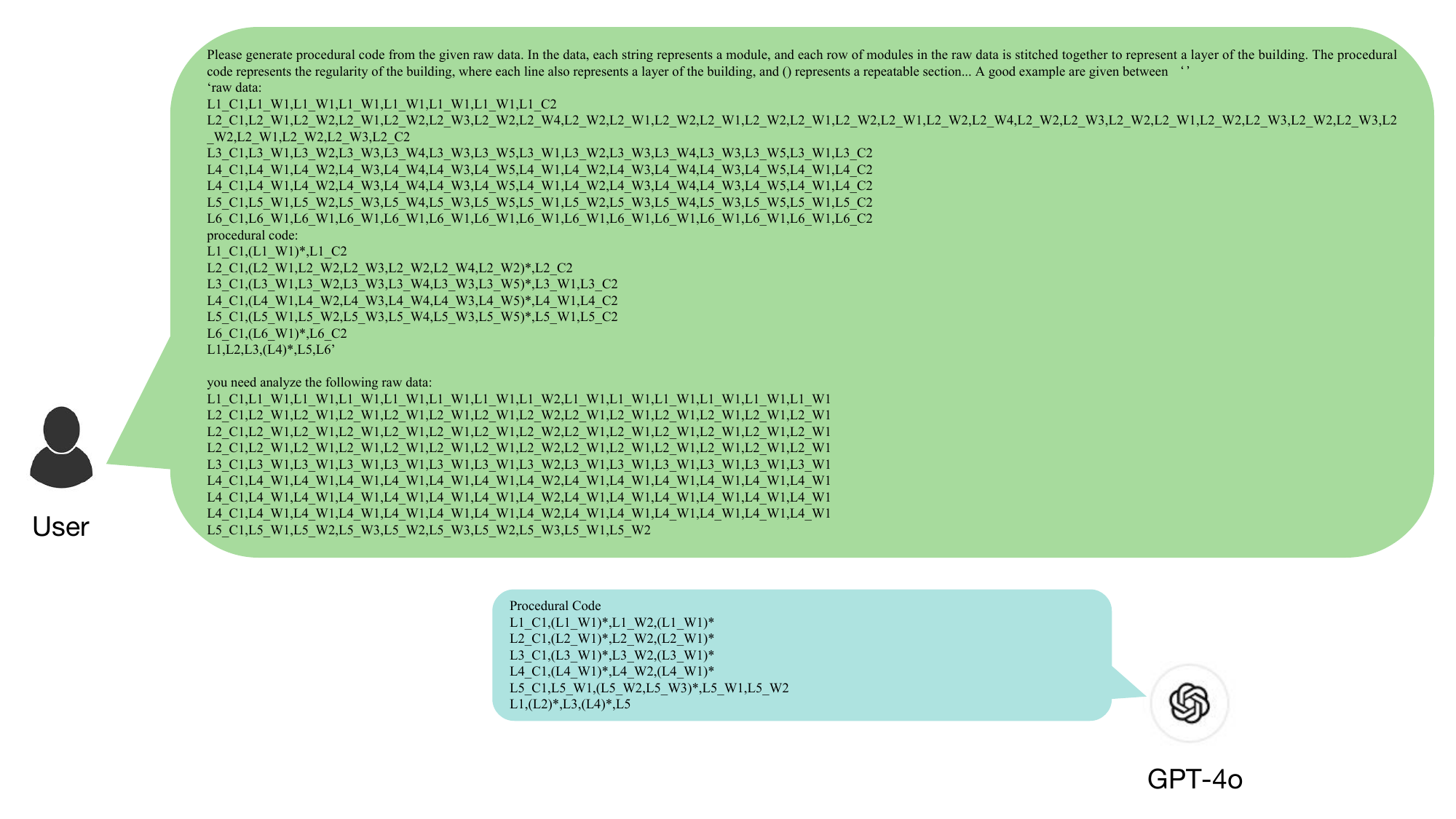}
\end{center}
\caption{\textbf{An example of the prompt used to obtain regular procedural code.} GPT-4o~\cite{hurst2024gpt} takes the raw data, one or more examples as well as descriptions of procedural code as input and summarizes the regular procedural code as output.}
\label{fig:prompt}
% \vspace{-2pt}
\end{figure*}

To illustrate the process of obtaining regular procedural code from raw data, we include an example of the prompt used in our framework in Figure~\ref{fig:prompt}. The goal is to summarize repetitive and scalable structures within raw data and represent them concisely using regular expressions of procedural code. The raw data represents the configuration of a multi-layered building with modular patterns. For instance, a single row might include repetitive modules like \emph{L1\_W1}. Learning from one or more pairs of raw data and regular procedural codes, GPT-4o~\cite{hurst2024gpt} could transforms raw data into a regularized procedural representation. We transform verbose raw data into a structured, succinct procedural summary that distills the input's intrinsic regularities while maintaining human interpretability.

\section{More Qualitative Results}
\label{sec:qualitative_results}
\begin{figure*}[t]
\begin{center}
\includegraphics[width=1.0\linewidth]{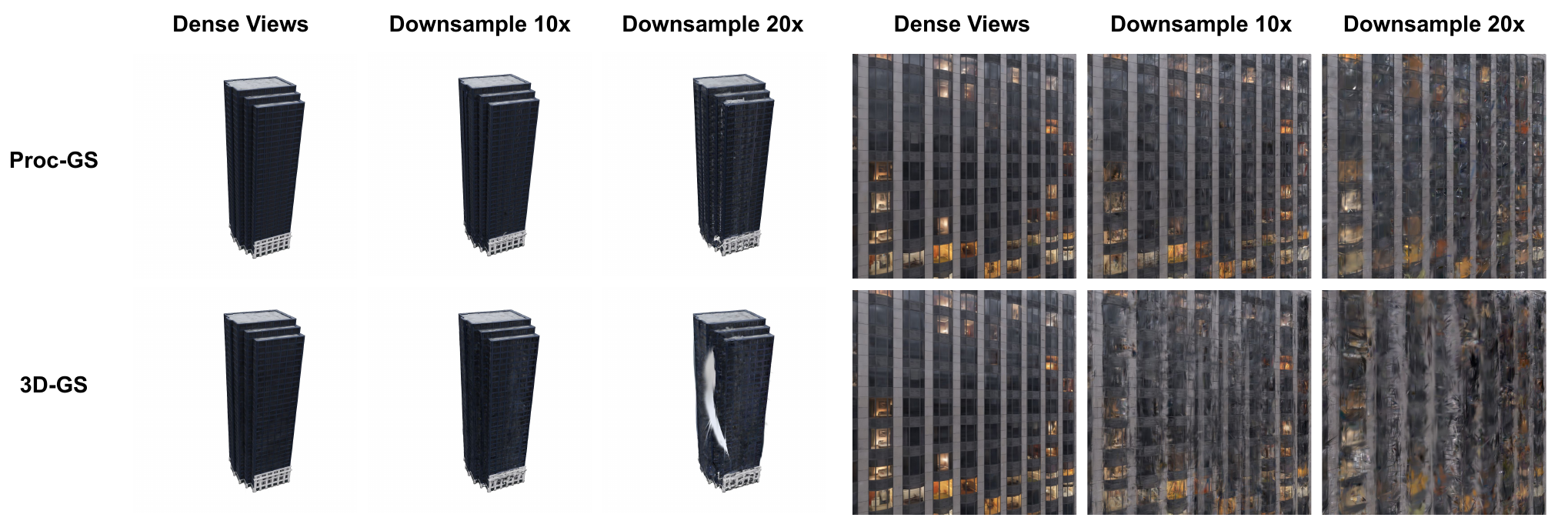}
\end{center}
\caption{\textbf{Sparse View qualitative results.} Proc-GS demonstrates significant robustness when reduces the number of training views, in contrast to 3D-GS~\cite{kerbl20233d}, which exhibits a pronounced susceptibility to numerous artifacts under similar conditions. This difference is attributed to the superior data efficiency inherent in our procedural code design, affirming its effectiveness in optimizing performance even with limited data inputs.}
\label{fig:sparse_view}
% \vspace{-2pt}
\end{figure*}

\begin{figure*}[t]
\begin{center}
\includegraphics[width=1.0\linewidth]{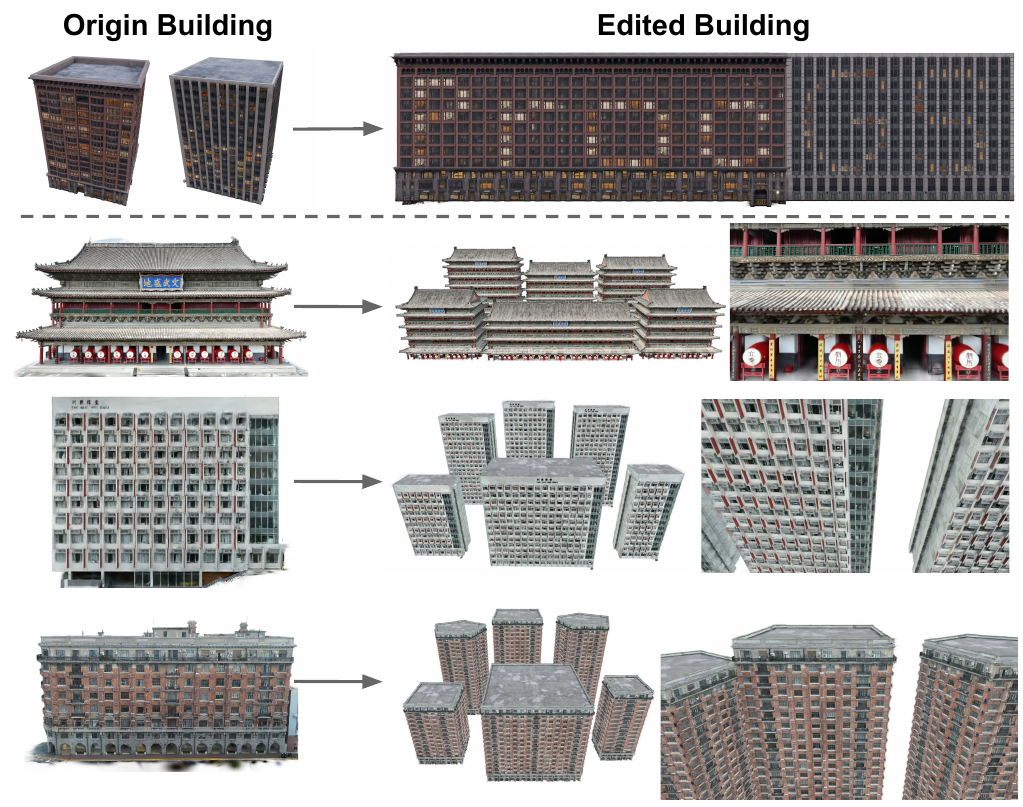}
\end{center}
\caption{\textbf{Building Editing}. (1) The upper part shows synthetic data results, where we arranged variance assets to spell our method's name, emphasizing its high controllability. (2) The lower part presents three editing results from the real-world scene.}
\label{fig:qualitative_results}
% \vspace{-2pt}
\end{figure*}

\subsection{Sparse View}\label{d.1}

Figure~\ref{fig:sparse_view} shows the sparse view qualitative results. Unlike 3D-GS~\cite{kerbl20233d}, which suffers from significant artifacts when reducing training views, Proc-GS exhibits remarkable robustness. The proposed design of shared base assets enables a natural data augmentation mechanism, where base assets are dynamically influenced by all instances throughout the training process. This characteristic significantly enhances our ability to extract base assets from sparse image sets, thereby substantially lowering the overall data collection expenses.

\subsection{Building Editing}\label{d.2}
In Figure~\ref{fig:qualitative_results}, we provide three building editing results from the real-world scene. We further showcase an intriguing building editing demo, where by manipulating variance assets, we precisely spelled out our method's name on the building facade, thereby illustrating the remarkable controllability of our approach. 

% We also provide \textbf{video demos} of building editing and city generation, which we strongly recommend viewing. In the video file named `city\_example.mp4', we show immersive traverse of a procedurally generated city from our assets constructed from the \emph{MatrixBuilding} dataset. In the video file named `building\_editing.mp4', we show the original building reconstructed from data collected in the real world, the base assets from our asset acquisition process, and the edited building with enlarged heights and widths.

% WARNING: do not forget to delete the supplementary pages from your submission 
% \input{sec/X_suppl}

\end{document}